\documentclass[11pt]{article}

\usepackage[final]{acl}
\usepackage{times}
\usepackage{latexsym}
\usepackage{amsmath}
\usepackage{amssymb}
\usepackage{booktabs}
\usepackage{tabularx}
\usepackage{placeins}
\usepackage{float}
\usepackage{graphicx}
\usepackage[T1]{fontenc}
\usepackage[utf8]{inputenc}
\usepackage{microtype}
\usepackage{inconsolata}

\title{RoTA: Role-Preserving Table Alignment for Frozen LLMs}

\author{
    Zixuan Zeng$^{1}$, Chenjie Qiu$^{1,2}$, Yuyan Wang$^{1}$, Zijian Li$^{3,4}$, Ruichu Cai$^{1,5}$, Boyan Xu$^{1,5}$\thanks{Corresponding author} \\
    $^{1}$School of Computer Science, Guangdong University of Technology \\
    $^{2}$Singapore Management University, Singapore \quad
    $^{3}$Carnegie Mellon University, Pittsburgh, USA\\
    $^{4}$Mohamed bin Zayed University of Artificial Intelligence \\
    $^{5}$Guangdong Provincial Key Laboratory of Media Convergence and Digital Creativity \\
    \texttt{zengzixuan44@mails.gdut.edu.cn} \quad \texttt{yausankit@gmail.com} \\
    \texttt{wangyuyan@mails.gdut.edu.cn} \quad
    \texttt{\{leizigin,cairuichu,hpakyim\}@gmail.com}
}

\begin{document}
\maketitle

\begin{abstract}
Large language models (LLMs) are increasingly used for table question answering, but serialized inputs can obscure the cell, row, and column roles that make table values meaningful. Existing table-as-modality methods encode table structure explicitly, yet often compress heterogeneous structural representations into a single table-level token, potentially causing structural role collapse. We propose \textbf{RoTA}, a compact \textbf{Ro}le-preserving \textbf{T}able \textbf{A}lignment interface for frozen LLMs. RoTA separately pools cell, row, and column representations from a hypergraph table encoder, maps them through role-specific projectors, and injects three structural soft prompt tokens. Across five table QA benchmarks, RoTA consistently outperforms the reproduced TAMO baseline, with gains of 2.62--5.50 accuracy points on four accuracy-based datasets and 2.00 BLEU points on FeTaQA. Shared-projector, role-ablation, and text-side perturbation experiments further support the contributions of role separation and role-specific alignment. Overall, preserving structural roles provides a simple and effective interface for table-to-LLM alignment. Our implementation will be open-sourced at \url{https://github.com/DMIRLAB-Group/RoTA}.
\end{abstract}

\section{Introduction}
A table cell rarely has a stable interpretation in isolation.
The value ``26'', for example, can denote an age, a score, a rank, or a year, depending on the row that identifies the record and the column that defines the attribute.
Table question answering (table QA) therefore requires models to interpret surface strings in light of their structural roles: cells provide local evidence, rows group records, and columns specify attributes \citep{tabert,tapas,turl}.
Most LLM-based table QA pipelines serialize tables into token sequences.
Although serialization preserves surface tokens, it encodes row membership, column membership, and cell identity only through textual cues such as delimiters, repeated headers, and field order.
Questions involving structural constraints therefore require models to reconstruct this layout from text, a key challenge for LLM-based table QA \citep{zhang2024table_reasoning_survey,tableformer}.

Prior work improves serialized-table reasoning through prompting, decomposition, tools, program execution, or instruction tuning \citep{dater,chainoftable,structgpt,binder,tablellama}, while table-aware encoders model structure explicitly \citep{tabert,tapas,tapex,turl,hytrel} (see Appendix~\ref{sec:related-work}).
TAMO globally pools cell, row, and column states into a single token and injects it into a frozen LLM \citep{tamo}.
The table encoder determines how structural relations are represented, while the alignment interface determines how these representations are presented to the LLM.
This raises a distinct question: should typed states remain separate at the encoder--LLM boundary?

Global pooling compresses heterogeneous cell, row, and column states into one table-level summary, removing explicit role separation from the alignment interface.
We call this interface-level loss \emph{structural role collapse} and hypothesize that preserving role distinctions during compression provides a more informative alignment interface.

We propose \textbf{RoTA}, a compact \textbf{Ro}le-preserving \textbf{T}able \textbf{A}lignment interface for frozen LLM table QA.
RoTA tests this hypothesis by separately summarizing cell, row, and column states and mapping the resulting summaries through role-specific projectors into three soft-prompt tokens.
Each token therefore exposes one structural role explicitly at the alignment boundary.

We evaluate RoTA on StructQA, WikiTQ, HiTab, WikiSQL, and FeTaQA.
Compared with our reproduction of TAMO, RoTA improves accuracy by 5.50, 3.43, 3.92, and 2.62 points on StructQA, WikiTQ, HiTab, and WikiSQL, respectively, and BLEU by 2.00 points on FeTaQA.
The shared-projector variant improves over global pooling, and role-specific projectors provide further gains; role-level ablations show that all three roles are useful.
Text-side permutation diagnostics further show that RoTA improves prediction consistency under row and column reorderings of the serialized table.

In summary, we identify structural role collapse as a limitation of globally pooled table-to-LLM alignment, introduce RoTA as a compact role-preserving interface, and provide evidence from benchmark results, role ablations, and text-side permutation diagnostics.

\section{Method}

\begin{figure*}[t]
\centering
\includegraphics[width=0.95\linewidth]{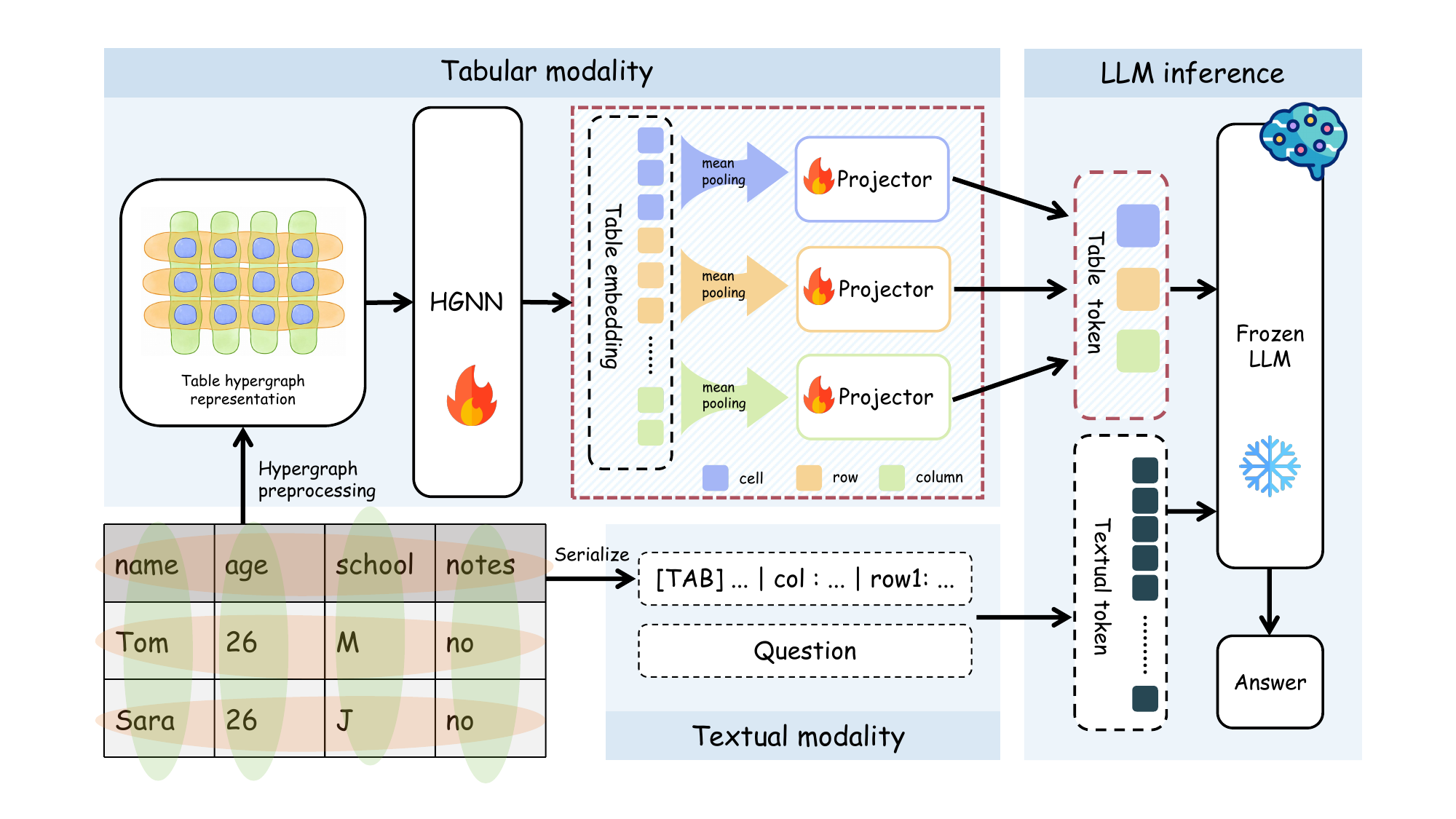}
\caption{Overall architecture of RoTA. The table is encoded as a hypergraph, separated into cell, row, and column roles, and injected as three structural soft prompt tokens.}
\label{fig:method}
\end{figure*}

\subsection{Problem Definition}

We study table question answering in a table-as-modality setting \citep{tamo}.
Each example is a triplet \((T,Q,A)\), where \(T\) is a pre-structured table, \(Q\) is a natural-language question, and \(A\) is the target answer sequence.
A standard LLM-based solution serializes \(T\) into a textual sequence \(S(T)\) and feeds the task instruction \(\mathcal{I}\), serialized table, and question into a decoder-only LLM.
We denote the resulting text-side embeddings as
\begin{equation}
X_{\mathrm{text}}
=
\operatorname{Emb}_{\mathrm{LLM}}([\mathcal{I};S(T);Q]).
\end{equation}

This textual stream provides lexical content but exposes row and column relations only through one-dimensional delimiters.

\subsection{Hypergraph Table Encoder}

RoTA follows TAMO and uses a hypergraph table encoder to obtain structural table representations \citep{tamo} (Figure~\ref{fig:method}).
The encoder architecture and training configuration are held constant across all compared variants, so the comparison focuses on how its outputs are aligned to the frozen LLM.

For a table \(T\), we construct a hypergraph \(\mathcal{G}=(\mathcal{V},\mathcal{E})\) (Appendix~\ref{sec:dataset-preprocessing}).
Cell-value nodes \(v\in\mathcal{V}\) contain table-cell text, while row and column groups are represented as hyperedges \(e\in\mathcal{E}\) connecting the cells in the corresponding row or column.
This construction provides explicit row-column membership information instead of requiring the LLM to infer such relations only from delimiters in \(S(T)\).

Following prior hypergraph table encoders \citep{allset,settransformer,hytrel,tamo}, the encoder performs node--hyperedge message passing with set-attention multiset operators.
After \(L\) layers, we collect cell-node states \(H_{\mathrm{cell}}\), row-hyperedge states \(H_{\mathrm{row}}\), and column-hyperedge states \(H_{\mathrm{col}}\) for downstream alignment.
RoTA pools and projects these role-specific states into structural prompts for the LLM.

\subsection{Role-Preserving Table Alignment}
The modality interface determines how table-encoder outputs enter the LLM embedding space.
A coarse interface pools all node and hyperedge states into a single table-structure token, which is compact but mixes structural roles before the LLM can use them.
RoTA instead treats each structural role as an alignment unit: cell states encode local value evidence, row hyperedges summarize record memberships, and column hyperedges summarize attribute memberships.
It keeps these groups separate until projection into the LLM space.

For each role \(r\in\{\mathrm{cell},\mathrm{row},\mathrm{col}\}\), RoTA first applies role-wise mean pooling in the hypergraph hidden space (see Appendix~\ref{sec:question-conditioned-aggregation} for question-conditioned alternatives):
\begin{equation}
h_r=\operatorname{MeanPool}(H_r).
\end{equation}
It then applies a role-specific projector:
\begin{equation}
z_r=P_r(h_r),
\quad
r\in\{\mathrm{cell},\mathrm{row},\mathrm{col}\},
\end{equation}
where \(P_{\mathrm{cell}}\), \(P_{\mathrm{row}}\), and \(P_{\mathrm{col}}\) are independent trainable projectors.
The structural prompt is
\begin{equation}
Z_T=[z_{\mathrm{cell}};z_{\mathrm{row}};z_{\mathrm{col}}]\in\mathbb{R}^{3\times d_l},
\end{equation}
where \(d_l\) is the LLM hidden dimension.

RoTA-shared uses the same three role-wise pooled vectors with a shared projector \(P\), while the coarse baseline globally pools all table representations before projection.

\subsection{Table-Text Interface and Training}

RoTA injects \(Z_T\) through an autoregressive soft-prompt interface.
The final input sequence is
\begin{equation}
\label{eq:input}
X=[x_{\mathrm{BOS}};Z_T;X_{\mathrm{text}}],
\end{equation}
where the structural tokens precede the textual stream. RoTA is trained as a lightweight table interface on top of a frozen LLM.
The trainable parameters are \(\Theta=\{\theta_{\mathrm{HG}},\theta_{\mathrm{proj}}\}\), while \(\theta_{\mathrm{LLM}}\) remains fixed.
For a training set \(\mathcal{D}\), RoTA minimizes
\begin{equation}
\vcenter{\hbox{\resizebox{0.92\columnwidth}{!}{$\displaystyle
\mathcal{L}(\Theta)
=
-\sum_{(T,Q,A)\in\mathcal{D}}\sum_{i=1}^{n}
\log p_{\theta_{\mathrm{LLM}}}
\left(a_i \mid X_\Theta(T,Q), a_{<i}\right),
$}}}
\end{equation}
where \(X_\Theta(T,Q)\) denotes the input sequence in Eq.~\ref{eq:input}, constructed with the question-independent structural prompt \(Z_T(T;\Theta)\).
Because gradients do not update the LLM, adaptation is concentrated in the table encoder and alignment projectors.

\section{Experiments}

\subsection{Experimental Setup}

\textbf{Datasets \& Metrics.}
We use five table QA benchmarks: StructQA, WikiTQ, HiTab, WikiSQL, and FeTaQA (Table~\ref{tab:appendix-datasets}).
StructQA is the primary structure-sensitive benchmark for diagnostics \citep{tamo}.
WikiTQ covers open-domain web tables \citep{pasupat2015wikitablequestions}, while HiTab targets hierarchical tables \citep{cheng2022hitab}.
WikiSQL evaluates question answering over tables through executable SQL queries \citep{zhong2017seq2sql}, whereas FeTaQA focuses on free-form answers grounded in tabular evidence \citep{nan2022fetaqa}.
We report denotation accuracy after answer normalization for StructQA, WikiTQ, HiTab, and WikiSQL, and BLEU for FeTaQA.

\textbf{Competing Methods.}
The main baseline is TAMO \citep{tamo}, which obtains one coarse table token through global pooling and a shared projector.
We reproduce TAMO with the same LLaMA2 backbone, hypergraph encoder, data splits, and training objective as RoTA.
RoTA-shared preserves cell, row, and column roles with a shared projector, while RoTA uses role-specific projectors.

\textbf{Implementation Details.}
All main experiments use a frozen LLaMA2 backbone and train only the table-side encoder and alignment modules \citep{touvron2023llama2}.
Other settings are fixed across methods to isolate the alignment interface.
Further details are in Appendix~\ref{sec:implementation-details} (Table~\ref{tab:appendix-hyperparams}).

\subsection{Main Results}

Table~\ref{tab:main-results} compares inference-only LLM prompting, TAMO, RoTA-shared, and RoTA on five benchmarks.

\begin{table}[t]
\centering
\small
\resizebox{\linewidth}{!}{%
\begin{tabular}{lrrrrr}
\toprule
Model & StructQA & WikiTQ & HiTab & WikiSQL & FeTaQA\\
\midrule
Inference only & 8.60 & 14.50 & 7.77 & 21.77 & 20.13\\
TAMO & 59.87 & 37.06 & 48.86 & 77.18 & 35.19\\
RoTA-shared & 63.20 & 38.45 & 51.01 & 79.18 & 36.41\\
RoTA & \textbf{65.37} & \textbf{40.49} & \textbf{52.78} & \textbf{79.80} & \textbf{37.19}\\
\midrule
\(\triangle\) TAMO & \(\uparrow\) 5.50 & \(\uparrow\) 3.43 & \(\uparrow\) 3.92 & \(\uparrow\) 2.62 & \(\uparrow\) 2.00\\
\bottomrule
\end{tabular}
}
\caption{Main results on table QA benchmarks. The metric is denotation accuracy except for FeTaQA, which uses BLEU.}
\label{tab:main-results}
\end{table}
RoTA outperforms TAMO by +5.50, +3.43, +3.92, and +2.62 accuracy points on StructQA, WikiTQ, HiTab, and WikiSQL, respectively, and by +2.00 BLEU points on FeTaQA.
RoTA-shared also improves over TAMO on StructQA by +3.33, showing that preserving cell, row, and column roles is beneficial even with a shared projector.
RoTA further improves over RoTA-shared by +2.17, suggesting that role-specific projectors provide a better alignment interface for heterogeneous table roles.
RoTA-shared also outperforms TAMO on WikiSQL and FeTaQA by +2.00 accuracy points and +1.22 BLEU points, respectively, while RoTA exceeds RoTA-shared by +0.62 and +0.78 points on the two benchmarks.
RoTA also outperforms TAMO with frozen Mistral-7B and Qwen3-8B backbones on both StructQA and WikiTQ; detailed results are provided in Appendix~\ref{sec:backbone-results}.

\subsection{Token Budget and Role-Assignment Control}
\label{sec:role-weighting-token-controls}

We compare RoTA-shared with controls that vary role weighting, token count, and group assignment while using a shared projector and the same trainable-parameter budget.
The \textit{role-balanced one-token} control averages the cell, row, and column means before projection.
The \textit{three-global-token} control repeats TAMO's projected global summary three times.
The \textit{random-role three-token} control randomly repartitions encoder outputs into groups matching the original role sizes.

Table~\ref{tab:alignment-controls} shows that RoTA-shared outperforms the role-balanced one-token and three-global-token controls by 3.33 and 2.13 accuracy points on StructQA, and by 0.28 and 0.37 points on WikiTQ, respectively.
These comparisons support presenting distinct role summaries, with a clearer advantage on StructQA.
Random role assignment yields 58.33 accuracy on StructQA, 4.87 points below RoTA-shared.
Together with the role ablations in Section~\ref{sec:role-ablation}, these results support organizing the structural prompt by cell, row, and column roles.

\begin{table}[H]
\centering
\small
\setlength{\tabcolsep}{4pt}
\begin{tabular}{lrr}
\toprule
Method & StructQA & WikiTQ \\
\midrule
TAMO & 59.87 & 37.06 \\
role-balanced one-token & 59.87 & 38.17 \\
three-global-token & 61.07 & 38.08 \\
random-role three-token & 58.33 & -- \\
RoTA-shared & \textbf{63.20} & \textbf{38.45} \\
\bottomrule
\end{tabular}
\caption{Shared-projector controls with equal trainable-parameter budgets. Results are denotation accuracy. The random-role control is evaluated on StructQA.}
\label{tab:alignment-controls}
\end{table}

\subsection{Parameter-Matched Control}

Compared with TAMO and RoTA-shared, the full RoTA model introduces two additional role-specific projectors. To control for the resulting increase in trainable parameters, we construct RoTA-shared+MLP, which retains the three role tokens and shared-projector design of RoTA-shared while adding a residual MLP shared across all three roles. RoTA-shared+MLP has exactly the same number of trainable parameters as RoTA but does not use role-specific projectors.

As shown in Table~\ref{tab:parameter-matched-control}, adding the shared residual MLP improves RoTA-shared by 1.27 points on StructQA and 0.32 points on WikiTQ. With exactly the same number of trainable parameters, RoTA further outperforms RoTA-shared+MLP by 0.90 and 1.72 points, respectively. These results provide evidence that role-specific projectors contribute beyond the increase in model capacity.

\begin{table}[H]
\centering
\small
\setlength{\tabcolsep}{4pt}
\begin{tabular}{lrrr}
\toprule
Model & Params. & StructQA & WikiTQ \\
\midrule
TAMO & 24.15M & 59.87 & 37.06 \\
RoTA-shared & 24.15M & 63.20 & 38.45 \\
RoTA-shared+MLP & 32.55M & 64.47 & 38.77 \\
RoTA & 32.55M & \textbf{65.37} & \textbf{40.49} \\
\bottomrule
\end{tabular}
\caption{Parameter-matched comparison on StructQA and WikiTQ. Results are denotation accuracy.}
\label{tab:parameter-matched-control}
\end{table}

\subsection{Ablation Study}
\label{sec:role-ablation}

We examine the contribution of each structural role on StructQA by masking one role or retaining only a single role (Table~\ref{tab:ablation}).

\begin{table}[!htbp]
\centering
\small
\setlength{\tabcolsep}{3.5pt}
\begin{tabular}{lcccrr}
\toprule
Variant & C & R & Col & Acc. ($\mu \pm \sigma$) & Drop \\
\midrule
RoTA & \checkmark & \checkmark & \checkmark & $\mathbf{65.34} \pm 0.45$ & 0.00 \\
mask cell & -- & \checkmark & \checkmark & $63.42 \pm 1.00$ & $-1.92$ \\
mask row & \checkmark & -- & \checkmark & $62.42 \pm 0.49$ & $-2.92$ \\
mask column & \checkmark & \checkmark & -- & $61.49 \pm 1.86$ & $-3.85$ \\
only cell & \checkmark & -- & -- & $59.40 \pm 0.75$ & $-5.94$ \\
only row & -- & \checkmark & -- & $60.35 \pm 1.34$ & $-4.99$ \\
only column & -- & -- & \checkmark & $61.15 \pm 0.55$ & $-4.19$ \\
\bottomrule
\end{tabular}
\caption{Role ablations on StructQA. C/R/Col denote cell, row, and column. Drop is computed relative to the complete RoTA model.}
\label{tab:ablation}
\end{table}

Masking the cell, row, or column token reduces accuracy by 1.92, 2.92, and 3.85 points, respectively.
Each single-role variant also achieves lower mean accuracy than the complete RoTA model, with drops ranging from 4.19 to 5.94 points.
These results support that the cell, row, and column tokens each contribute useful structural information and that jointly preserving all three roles benefits model performance.

\subsection{Permutation Robustness}

Serialized table prompts can be sensitive to row and column order even when table semantics should remain stable \citep{tableformer,hytrel}.
Following TAMO-style robustness evaluation \citep{tamo}, we perturb only the serialized table while keeping the graph-side structural representation fixed (Appendix~\ref{sec:perturbation-details}).
This diagnostic evaluates sensitivity to textual serialization order with a fixed structural representation.
We report perturbed accuracy and consistency, where consistency measures whether the prediction remains unchanged after permutation.

Table~\ref{tab:robustness} shows that RoTA improves accuracy over TAMO by +5.50, +5.33, +4.07, and +4.00 points on the original, row-, column-, and row+column-perturbed tables, respectively.
It also improves prediction consistency by +3.47, +5.00, and +4.40 points under row, column, and joint permutations, respectively.
These consistency gains indicate more stable predictions across different table linearizations.

\begin{table}[H]
\centering
\small
\resizebox{\linewidth}{!}{%
\begin{tabular}{lccccccc}
\toprule
Method & Orig. & \multicolumn{2}{c}{Row} & \multicolumn{2}{c}{Col.} & \multicolumn{2}{c}{Row+Col.} \\
\cmidrule(lr){3-4}\cmidrule(lr){5-6}\cmidrule(lr){7-8}
 & Acc. & Acc. & Cons. & Acc. & Cons. & Acc. & Cons. \\
\midrule
TAMO & 59.87 & 54.80 & 58.33 & 56.20 & 57.67 & 53.27 & 50.60 \\
RoTA & \textbf{65.37} & \textbf{60.13} & \textbf{61.80} & \textbf{60.27} & \textbf{62.67} & \textbf{57.27} & \textbf{55.00} \\
\bottomrule
\end{tabular}
}
\caption{Text-side robustness on StructQA. Orig. is the original table; Row/Col./Row+Col. are perturbations. Acc. denotes accuracy, Cons. consistency.}
\label{tab:robustness}
\end{table}

\subsection{Case Study}

To examine whether role-preserving alignment helps locate question-relevant table evidence, we conduct a gradient-based visual case study. Following PromptBench~\citep{zhu2023promptbench}, we compute the gradient norm of the gold-answer generation loss with respect to the embedding at each input position as an estimate of input importance. TAMO and RoTA use the same table, serialized input, question, tokenizer, and gold-answer objective. We randomly select one WikiTableQuestions test example from cases for which TAMO produces an incorrect answer while RoTA produces the correct answer.

\begin{figure}[!t]
\centering
\includegraphics[width=0.98\columnwidth]{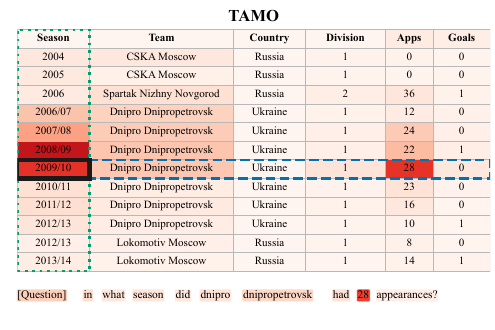}
\vspace{2pt}

\includegraphics[width=0.98\columnwidth]{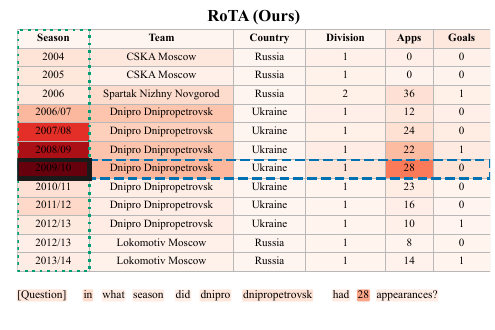}
\caption{Input-embedding gradient attribution for the WikiTableQuestions question ``In what season did Dnipro Dnipropetrovsk have 28 appearances?'' TAMO incorrectly predicts 2008/09, whereas RoTA correctly predicts the gold answer 2009/10. The blue dashed, green dotted, and solid black borders denote the target row, target column, and gold cell, respectively. Darker colors indicate greater gold-answer-conditioned input-embedding gradient attribution under a shared, unclipped display scale.}
\label{fig:wtq-gradient-case-study}
\end{figure}

Figure~\ref{fig:wtq-gradient-case-study} shows that TAMO emphasizes the adjacent distractor, while RoTA assigns the highest importance to the gold cell. Gold-cell attribution rises from 10.8\% to 14.7\% and target-column attribution from 20.0\% to 24.5\%; target-row attribution remains similar (28.0\% vs.\ 28.3\%). This case suggests that RoTA better links \texttt{Apps = 28} to \texttt{Season = 2009/10} within the same row using row and column context.

\section{Conclusion}

We introduced \textbf{RoTA}, a role-preserving interface for frozen LLM table QA.
The structural prompt remains fixed at three tokens across table sizes, providing a compact interface for exposing cell, row, and column summaries.
RoTA improves over the reproduced TAMO baseline across five benchmarks.
Controlled comparisons and role ablations support the contributions of role separation and role-specific alignment, while text-side permutation diagnostics show higher prediction consistency on StructQA.
These results highlight the value of preserving structural roles at the encoder--LLM boundary.

\section*{Limitations}

First, our main comparison focuses on TAMO as the primary baseline.
Broader comparisons with other LLM-based table understanding systems would strengthen the empirical positioning.

Second, the main five-benchmark evaluation uses a frozen LLaMA2 backbone to isolate the effect of table-to-LLM alignment \citep{touvron2023llama2}.
Appendix~\ref{sec:backbone-results} reports controlled results with frozen Mistral-7B and Qwen3-8B backbones on StructQA and WikiTQ, but we have not evaluated RoTA with parameter-efficient adaptation or full supervised fine-tuning \citep{jiang2023mistral7b,yang2025qwen3technicalreport,hu2022lora}.

Third, RoTA assumes pre-structured tables and relies on role metadata from preprocessing.
Errors in table parsing, row/column recognition, or role assignment may reduce the quality of the structural prompts.
For hierarchical tables such as HiTab, cell, row, and column roles provide only a coarse decomposition and may not fully capture multi-level header semantics \citep{cheng2022hitab}.
Representing each role with a single question-independent token may also limit fine-grained localization when different questions concern different cells, rows, or columns.
More parameter-efficient question-conditioned aggregation or alignment mechanisms may provide further benefits for questions requiring precise evidence selection.

\section*{Acknowledgments}
This research was supported in part by National Science and Technology Major Project (2021ZD0111502), Natural Science Foundation of China (U24A20233, 62406078), Guangdong Provincial Key Laboratory of Media Convergence and Digital Creativity (2026B1212020003) and CCF-DiDi GAIA Collaborative Research Funds (CCF-DiDi GAIA 202521).

\bibliography{references}

\appendix
\section{Related Work}
\label{sec:related-work}

\subsection{LLM-based Table Reasoning}

With the rapid development of large language models, LLM-based table reasoning has become a major paradigm for table question answering and related structured-data tasks \citep{zhang2024table_reasoning_survey}. Most existing methods first linearize a table into Markdown, HTML, CSV-like text, or natural-language records, and then improve the LLM's reasoning ability through either parameter adaptation or prompt engineering. Fine-tuning-based methods adapt LLMs to tabular or structured-data tasks using supervised fine-tuning, instruction tuning, LoRA-style adaptation, or additional table-oriented pretraining \citep{structlm,protrix,sarkar2023unified_table_pretraining,llasa,hegta,tabglm,long2025semantic_gap_nl2sql,hyperg}. TableLLaMA is a representative system in this direction: it builds an open generalist model for tables by instruction-tuning open LLMs on a broad collection of tabular tasks \citep{tablellama}. Table structure is also incorporated through special-token encoding for serialization and 2D LoRA for modeling cell positions \citep{he-etal-2025-tablelora}. These methods improve task performance, but they usually require LLM-side adaptation, multi-stage training, or large-scale table-oriented pretraining. TAMO is closest to our setting because it establishes table-as-modality as an effective framework for frozen LLM table QA: it represents a table as a hypergraph, encodes the structure with a table encoder, and injects the resulting table representation into a frozen LLM together with the linearized textual input \citep{tamo}. However, its alignment stage compresses heterogeneous cell, row, and column representations before exposing them to the LLM. RoTA follows the same table-as-modality setting but studies a focused alignment bottleneck: whether structural roles should be preserved before projection into the LLM embedding space. Prompt-engineering methods instead keep the LLM largely fixed and use carefully designed prompts, decomposition strategies, symbolic execution, or verification to improve table reasoning over serialized tables \citep{dater,chainoftable,structgpt,binder,lever}. These methods strengthen the reasoning process, but table structure still mainly enters the LLM as a one-dimensional sequence or as operations over that sequence, leaving row-column relations and structural roles only implicit.

\subsection{Table Encoder}

A broad line of work explores how to encode, represent, and understand tabular data. Early table encoders learn table-aware representations by adding structural information to pretrained language models. TaBERT jointly models textual contexts and table content \citep{tabert}; TAPAS incorporates row, column, and rank embeddings into a BERT-style encoder for weakly supervised table parsing \citep{tapas}; and TAPEX uses executable supervision to pretrain an encoder-decoder model as a neural SQL executor \citep{tapex}. Other work studies structural attention, feature selection, and set- or graph-based table representation learning \citep{tableformer,tabnet,settransformer}. Hypergraph-based encoders are especially relevant because they can model high-order relations among table elements. HyTrel represents tables with hyperedges over cells, rows, columns, and the whole table \citep{hytrel}, while general hypergraph neural networks provide mechanisms for propagating information over such high-order structures \citep{hypergcn,allset}. These methods show that explicit table encoders can capture structural signals that are difficult to recover from plain text serialization. However, most of them are designed for encoder-only models, encoder-decoder models, or standalone tabular prediction pipelines, rather than as modality-injection interfaces for modern decoder-only LLMs. The hypergraph construction in RoTA follows prior hypergraph table representation work, while our contribution lies in how the resulting cell, row, and column representations are aligned to the frozen LLM.

\section{Experiments}

\subsection{Implementation Details}
\label{sec:implementation-details}

\textbf{The table encoder.} A pretrained RoBERTa-large model provides 1024-dimensional table-text features \citep{reimers2019sentencebert}, which are projected into a one-layer hypergraph transformer with hidden size 1024 and four attention heads. Role labels distinguish cell, row, and column outputs.

\textbf{LLMs.} The main five-benchmark evaluation uses frozen LLaMA2-7B with LLaMA2-style instruction prompts \citep{touvron2023llama2}. The serialized-table limit increases from 1024 tokens during training to 4096 at test time to reduce truncation. Loading, decoding, and checkpoint settings are listed in Table~\ref{tab:appendix-hyperparams}; Mistral-7B and Qwen3-8B settings are described in Appendix~\ref{sec:backbone-results}.

\textbf{Optimization.} Optimization settings and hardware are summarized in Table~\ref{tab:appendix-hyperparams}.

\textbf{Role-ablation protocol.} The ablations in Section~\ref{sec:role-ablation} use three random seeds: 7, 42, and 2026. We keep the three-token interface length fixed and replace each ablated role token with zeros. Table~\ref{tab:ablation} reports the mean accuracy and standard deviation across the three runs; drops are computed relative to the mean accuracy of the complete RoTA model.

\begin{table*}[!t]
\centering
\small
\setlength{\tabcolsep}{4pt}
\begin{tabularx}{\textwidth}{@{}>{\raggedright\arraybackslash}p{0.15\textwidth}>{\raggedright\arraybackslash}X>{\raggedright\arraybackslash}p{0.12\textwidth}>{\raggedright\arraybackslash}p{0.21\textwidth}@{}}
\toprule
Dataset & Split used by implementation & Evaluation split & Metric \\
\midrule
StructQA / StructProbe & train / validation / test, 60\% / 20\% / 20\% over 7,500 examples & test & denotation accuracy \\
WikiTQ & official train / validation / test split from preprocessing & test & denotation accuracy \\
HiTab & official train / validation / test split from preprocessing & test & denotation accuracy \\
WikiSQL & official train / validation / test split from preprocessing & test & denotation accuracy over executed answers \\
FeTaQA & official train / validation / test split from preprocessing & test & BLEU (SacreBLEU, lowercase) \\
\bottomrule
\end{tabularx}
\caption{Datasets, splits, and evaluation metrics. The implementation uses the same split names for all five datasets and selects the final checkpoint by validation loss before test evaluation.}
\label{tab:appendix-datasets}
\end{table*}

StructQA (\texttt{structprobe}) contains 7,500 QA pairs built from WikiTQ tables, split into 4,500 training, 1,500 validation, and 1,500 test examples. The splits contain 300, 100, and 100 tables, respectively, with 499 unique tables overall. Test tables are disjoint from training and validation; the latter two splits share one table and three duplicate QA pairs.

\begin{table*}[!t]
\centering
\small
\setlength{\tabcolsep}{4pt}
\begin{tabularx}{\textwidth}{@{}>{\raggedright\arraybackslash}p{0.20\textwidth}>{\raggedright\arraybackslash}X@{}}
\toprule
Category & Setting \\
\midrule
Backbone & frozen LLaMA2-7B, bfloat16 loading \\
Prompt format & LLaMA2-style instruction prompt \\
Optimizer & AdamW \citep{loshchilov2019adamw}, learning rate \(1\times10^{-5}\), weight decay \(0.05\), betas \((0.9, 0.95)\) \\
Training schedule & 10 epochs, 1 warmup epoch, half-cycle cosine decay, early stopping patience 3, gradient clipping 0.1 \\
Batching & batch size 4 with one optimizer step per mini-batch; evaluation batch size 8 \\
Input and decoding & maximum training text length 1024, maximum test text length 4096, maximum new tokens 128 \\
Hypergraph encoder & 1 layer, hidden size 1024, 4 attention heads, encoder dropout 0.1 \\
Structural alignment & TAMO uses one pooled token; RoTA-shared uses three role tokens with one shared projector; RoTA uses three role tokens with role-specific projectors \\
Random seed & 42 for training, evaluation, and perturbation generation unless explicitly overridden \\
Checkpoint selection & best validation loss checkpoint is reloaded for test evaluation \\
Hardware & one NVIDIA A800 GPU per run \\
\bottomrule
\end{tabularx}
\caption{Hyperparameters and implementation settings used by the main experiments.}
\label{tab:appendix-hyperparams}
\end{table*}

\subsection{Dataset Preprocessing}
\label{sec:dataset-preprocessing}

Each table is serialized as column headers followed by row-wise cell values and converted into a bipartite hypergraph. For StructQA, WikiTQ, WikiSQL, and FeTaQA, cell-value nodes connect to their corresponding row and column hyperedges. Row hyperedges are indexed by row position, and column hyperedges by header text. Text normalization allows repeated cell values or structural descriptors to share node or hyperedge representations within an example.

For HiTab, merged cells are expanded to fill their covered positions. Traversing the top and left header trees recovers the header paths for each data cell, which is connected to the row- and column-side header texts along these paths. For WikiSQL, SQL annotations are executed against typed table rows to obtain answer strings for denotation evaluation.

We cache the pretrained cell and hyperedge features, incidence matrices, and role labels as graph objects for reuse during training and evaluation across methods.

\subsection{Perturbation Details}
\label{sec:perturbation-details}

We construct row, column, and joint row--column permutations of StructQA. Row permutations shuffle the body rows. Column permutations keep the first column fixed to preserve the row identifier and shuffle the remaining columns; joint permutations apply both transformations.

Permutations are generated deterministically using seed 42 with an example-specific offset. Only the serialized table text is modified; questions, answers, and the original graph features remain unchanged. This diagnostic measures prediction consistency under different table serializations.

\FloatBarrier
\subsection{Question-Conditioned Aggregation}
\label{sec:question-conditioned-aggregation}

We compare question-conditioned attention pooling in the GNN representation space and the LLM embedding space. For each role, the question representation serves as the attention query over the corresponding role representations. The resulting summaries are injected into the same frozen LLM; the encoder, training objective, and evaluation protocol match RoTA.

Table~\ref{tab:question-conditioned-aggregation} shows mixed benefits from question-conditioned pooling. The GNN-space variant lowers accuracy, while the LLM-space variant gains 0.96 points but adds approximately 378M trainable parameters, compared with RoTA's 32.55M total. Mean pooling remains a compact design among the evaluated variants.

RoTA's question-independent structural tokens can also be reused across questions about the same table.

\begin{table}[H]
\centering
\small
\setlength{\tabcolsep}{2pt}
\begin{tabular}{@{}lrrr@{}}
\toprule
Method & Acc. & $\triangle$ & Add. Params. \\
\midrule
RoTA: mean pooling & 65.37 & -- & -- \\
RoTA+: GNN-space pooling & 64.87 & $-0.50$ & approx. +5M \\
RoTA+: LLM-space pooling & \textbf{66.33} & $+0.96$ & approx. +378M \\
\bottomrule
\end{tabular}
\caption{Comparison of mean pooling and question-conditioned aggregation on StructQA.}
\label{tab:question-conditioned-aggregation}
\end{table}

\FloatBarrier
\subsection{Results across Frozen LLM Backbones}
\label{sec:backbone-results}

We evaluate TAMO, RoTA-shared, and RoTA with frozen Mistral-7B and Qwen3-8B on StructQA and WikiTQ \citep{jiang2023mistral7b,yang2025qwen3technicalreport}. Within each backbone, data splits, serialization, the encoder, training objective, optimization, and evaluation remain fixed across methods. Backbone-specific changes are limited to the tokenizer, prompt template, and projection dimensions.

Across both backbones and datasets, RoTA-shared outperforms TAMO, and RoTA provides further gains (Table~\ref{tab:backbone-results}). These results show that the improvements extend beyond LLaMA2-7B.

\begin{table}[H]
\centering
\small
\setlength{\tabcolsep}{3pt}
\begin{tabular}{@{}lrrrr@{}}
\toprule
Method & \multicolumn{2}{c}{StructQA} & \multicolumn{2}{c}{WikiTQ} \\
\cmidrule(lr){2-3}\cmidrule(lr){4-5}
 & Acc. & $\triangle$ & Acc. & $\triangle$ \\
\midrule
\multicolumn{5}{@{}l}{\textit{Mistral-7B / Mistral-7B-Instruct}} \\
Inference only & 15.00 & -- & 19.00 & -- \\
TAMO & 70.00 & -- & 48.16 & -- \\
RoTA-shared & 71.13 & +1.13 & 50.18 & +2.02 \\
RoTA & \textbf{74.27} & +4.27 & \textbf{52.92} & +4.76 \\
\midrule
\multicolumn{5}{@{}l}{\textit{Qwen3-8B}} \\
Inference only & 2.11 & -- & 6.53 & -- \\
TAMO & 66.73 & -- & 56.42 & -- \\
RoTA-shared & 73.93 & +7.20 & 57.11 & +0.69 \\
RoTA & \textbf{80.07} & +13.34 & \textbf{59.41} & +2.99 \\
\bottomrule
\end{tabular}
\caption{Results with different frozen LLM backbones on StructQA and WikiTQ. All values are denotation accuracy, and each difference is computed relative to TAMO with the same backbone. The Mistral-family inference-only result uses Mistral-7B-Instruct, whereas TAMO, RoTA-shared, and RoTA use Mistral-7B.}
\label{tab:backbone-results}
\end{table}

\end{document}